\documentclass[letterpaper]{article} 
\usepackage{aaai2026}  
\usepackage{times}  
\usepackage{helvet}  
\usepackage{courier}  
\usepackage[hyphens]{url}  
\usepackage{graphicx} 
\usepackage{natbib}  
\usepackage{caption} 
\usepackage{algorithm}
\usepackage{algorithmic}

\usepackage{amsmath}
\usepackage{amsthm}
\usepackage{paralist}
\usepackage{dsfont}
\usepackage{xcolor} 
\usepackage{xspace}
\usepackage{amssymb}
\usepackage{subcaption}
\allowdisplaybreaks

\newcommand{\gregor}[1]{\textcolor{red}{\textbf{Gregor:} #1}}
\renewcommand{\gregor}[1]{#1}

\newtheorem{example}{Example}
\newtheorem{definition}{Definition}

\newif\ifextendedversion
\extendedversiontrue   

\usepackage{newfloat}
\usepackage{listings}
\DeclareCaptionStyle{ruled}{labelfont=normalfont,labelsep=colon,strut=off} 
\floatstyle{ruled}
\newfloat{listing}{tb}{lst}{}
\floatname{listing}{Listing}
\title{When both Grounding and not Grounding are Bad -- \\ A Partially Grounded Encoding of Planning into SAT}
\author {
    João Filipe\textsuperscript{\rm 1},
    Gregor Behnke\textsuperscript{\rm 1}
}
\affiliations {
    \textsuperscript{\rm 1}University of Amsterdam, 
    Institute for Logic Language and Computation, 
    The Netherlands\\
    j.sa@uva.nl, g.behnke@uva.nl
}

\usepackage{bibentry}

\begin{document}

\maketitle

\begin{abstract}
Classical planning problems are typically defined using lifted first-order representations, which offer compactness and generality. While most planners ground these representations to simplify reasoning, this can cause an exponential blowup in size. Recent approaches instead operate directly on the lifted level to avoid full grounding.

We explore a middle ground between fully lifted and fully grounded planning by introducing three SAT encodings that keep actions lifted while partially grounding predicates.
Unlike previous SAT encodings, which scale quadratically with plan length, our approach scales linearly, enabling better performance on longer plans.
Empirically, our best encoding outperforms the state of the art in length-optimal planning on hard-to-ground domains.
\end{abstract}


\section{Introduction}

Automated Planning aims to find a sequence of actions that
transforms an initial state into one satisfying a given goal.
While most research in the last decades focussed on solving planning problems that can be fully ground instantiated, the task of lifted planning - that is planning with actions described using a function-free first-order language - has made significant progress since the seminal work by \citeauthor{correa-et-al-icaps2020}~(\citeyear{correa-et-al-icaps2020}).
Using lifted planning, significantly larger problems than before could be tackled successfully.

Current methods for lifted planning mainly fall into two categories: ones based on search and ones based on translation to logic.
In this paper, we focus on the latter.

\citeauthor{shaik-vandepol-icaps2022}~(\citeyear{shaik-vandepol-icaps2022}) proposed a translation of lifted planning into Quantified Boolean Formulae (QBF), in which lifting is encoded using universal quantification by iterating over possible objects as arguments for predicates.
At the same time \citeauthor{hoeller-behnke-icaps2022}~(\citeyear{hoeller-behnke-icaps2022}) proposed an encoding into Boolean Satisfiability (SAT) -- called LiSAT -- which is until now the state-of-the-art for plan-length optimal planning, which is also optimal planning for unit-cost problems.
Similar to grounded SAT-based planners, it views the plan as a sequence of time steps and generates SAT formulae $\phi_\ell$ that are satisfiable iff there is a plan of length up to $\ell$.
To generate an optimal plan, it iterates over $\ell$ until a plan is found.
LiSAT expresses lifting using variables $a^t$ and $(v=o)^t$ denoting that action $a$ is executed at time $t$ and argument variable $v$ of that action is the object $o$, respectively.
To encode that preconditions of actions are met and that effects are applied, LiSAT uses the concept of causal links~\cite{penberthy-weld-kr1992}, linking actions having a fact as a precondition to actions that make that fact true.
As a consequence, the size of $\phi_\ell$ grows quadratically in $\ell$ -- which is not desirable.

We explore the middle ground between full grounding and full lifting for SAT-based planning.
For this, we introduce three encodings that require only a partial grounding of the planning problem.
In all three, actions are encoded fully lifted.
For the state description, i.e., the problem's predicates, we investigate both a full grounding approach, as well as two approaches that represent the current state using lifted representations that explicitly encode predicates and their argument objects.
For the latter two, we use lifted mutex groups \cite{helmert-aij2009}, allowing for a compact lifted encoding of some of the problem's predicates.
We present two ways of encoding objects in this case: with a one-hot encoding or with a binary encoding.

Although the full grounding approach is not competitive with other SAT-based approaches, our new partially grounded encodings outperform LiSAT in five out of nine benchmark domains.

\ifextendedversion
\else
An extended version is available~\cite{arxivextendedversion}.
\fi

\section{Preliminaries}

\paragraph{Lifted Planning}
We formalize lifted STRIPS planning problems~\cite{fikes-nilsson-aij1971} as a tuple: $\Pi = \langle \mathcal{O}, \mathcal{P}, \mathcal{A}, \mathcal{I}, \mathcal{G} \rangle$.
Here, $\mathcal{O}$ is a set of objects, $\mathcal{P}$ is a set of predicates symbols, $\mathcal{A}$ is a set of action symbols, $\mathcal{I}$ is the initial state and $\mathcal{G}$ is the goal condition.
Each object $o \in \mathcal{O}$ has a type $t$.
The set of all objects of a type is denoted as $\mathcal{O}^t$.
Similarly, for a variable $v$ we write $\mathcal{O}^v$ to denote all objects of $v$'s type.
Typically, types are defined via a type hierarchy, where one type $t_1$ is a subtype of another type $t_2$.
We flatten this hierarchy by adding all objects of $\mathcal{O}^{t_1}$ to $\mathcal{O}^{t_2}$.
We assume that the type hierarchy is a tree (which it always is in benchmark instances).
Each predicate symbol $p^n \in \mathcal{P}$ an arity $n$ and a tuple of typed variables: $p^n(v_1^{t_1}, \dots ,v_n^{t_n})$, where $v_i^{t_i}$ indicates that variable $v_i$ is of the type $t_i$.
We call a predicate together with concretely chosen arguments, which can either be variables or objects $p^n(a_1, \dots ,a_n)$ an \emph{atom}.
When all variables are replaced by objects of the respective type, the predicate is considered grounded, then called a fact.
Similarly to predicate symbols, each action symbol $a^n \in \mathcal{A}$ has an arity $n$, a tuple of typed variables and is considered grounded when all variables are replaced by objects of the respective type.
To describe the action's semantics, there are three functions: $prec$, $add$ and $del$.
They map actions to their preconditions, add effects, and delete effects, respectively, which are sets of atoms using the action's arguments as variables.
Lastly, $\mathcal{I}$ and $\mathcal{G}$ are sets of facts, representing the initial state and the goal condition.    

\begin{definition}
    (Applicability). A ground action $a(o_1,\dots,o_n)$ is applicable in state $s$ iff $prec(a(o_1,\dots,o_n)) \subseteq s$.
\end{definition}

\begin{definition}
    (State Transition). Given a state $s$ and an applicable ground action $a$, the transition function $\tau(s,a(o_1,\dots,o_n)) = (s \backslash del(a(o_1,\dots,o_n))) \cup add(a(o_1,\dots,o_n))$ gives the resulting state.
\end{definition}

\begin{definition}
    (Solution). A solution is a sequence of ground actions $\pi=(a_0(o_1^1,\dots,o_{n_1}^1),\dots,a_m(o_1^m,\dots,o_{n_m}^m))$ such that $a_{i}$ is applicable in $s_i$, $s_{i+1} = \tau (s_i, a_{i}(o_1^{i},\dots,o_{n_{i}}^{i}))$, $s_0 = \mathcal{I}$ and $\mathcal{G} \subseteq s_{m+1}$
\end{definition}

\begin{definition}
    (Optimal Length). A solution $\pi=(a_1,\dots,a_n)$ is length optimal if for any other solution $\pi ' =(b_1,\dots,b_m)$, it holds that $n \leq m$.
\end{definition}


\paragraph{Lifted Mutex Groups}
In planning problems, it is often the case, that not all syntactically valid states can actually be reached from $\mathcal I$.
Instead, certain invariants are true for every state reachable via application of actions.
One such type of invariants are mutex groups: they state that, out of a set of facts, at most one can be true in any reachable state.
For example, a package cannot simultaneously be in two different locations. Thus, facts such as $at(package1,lisbon)$ and $at(package1,amsterdam)$ would form a mutex group. 

Defining, computing, and using mutex groups as sets of facts is problematic in practice as there are often many mutex groups with complex structures.
Instead, one can restrict to lifted mutex groups (LMGs, \citeauthor{helmert-aij2009}~\citeyear{helmert-aij2009}).
A LMG is described using variables so that every ground instantiation of an LMG is a mutex group.
LMGs can be obtained without ground-instantiating all facts and actions~\cite{helmert-aij2009,fiser-aaai2020}, but these procedures usually do not find all LMGs, but only a (practically sufficient) subset of them.
For our experiments, we used the discovery method by \citeauthor{fiser-aaai2020}~\shortcite{fiser-aaai2020} based on Lifted Fact Altering Mutex Groups, which can generate a substantial amount of LMGs quickly.

We now define the syntax for LMGs, by defining candidate structures that could be LMGs.
\begin{definition}
    A lifted mutex group (LMG) candidate $\mathcal L$ is a triple $\langle \textit{fix},\textit{cnt}, P\rangle$ where:
    (1) \textit{fix} is a set of \emph{fixed variables},
    (2) \textit{cnt} is a set of \emph{counted variables} with $\textit{fix} \cap \textit{cnt} = \emptyset$, and
    (3) $P$ is a set of atoms, whose variables are from $\textit{fix} \cup \textit{cnt}$.
\end{definition}
Conceptually, each instantiation of the fixed variables $fix$ should yield a mutex group.
The facts in that mutex group are obtained by computing all ground-instantiations of the counted variables $cnt$.

For example, we may have $\mathcal L = \langle \{?p\}, \{?l\}, \{at(?p, ?l)\} \rangle$.
Each instantiation of the fixed variable $?p$ with an object $x$ will yield one mutex group, whose elements are the facts of the form $at(x,y)$ -- where $y$ is a possible object for the variable $?l$.

For the purposes of this paper, we often work with partially instantiated LMGs, which we call partially lifted mutex groups (PLMGs).
In them, the fixed variables have already been instantiated.

\begin{definition}
    A partially lifted mutex group (PLMG) candidate $\mathcal M$ is a pair $\langle cnt, P\rangle$ where:
     $cnt$ is a set of \emph{counted variables}, and
     $P$ is a set of atoms, with variables from $cnt$.\\
    We write $\mathfrak C(\mathcal M)$ to denote the counted variables of a PLMG and $\mathfrak L(\mathcal M)$ to denotes its atoms.
\end{definition}

In both LMGs and PLMGs, all variables $v$ are associated with a type $\mathcal O^v$.
Given a LMG and an instantiation of the fixed variables $I: fix \rightarrow O$, we obtain a PLMG by replacing all occurrences of fixed variables $v$ by $I(v)$.
A PLMG implies a single mutex group $\mathfrak M$ by computing all ground instantiations of its atoms w.r.t\ to the counted variables $cnt$.
We write this set of ground instantiated facts as $F(\mathcal M)$.

\begin{definition}
    A PLMG candidate $\mathcal{M}$ is a PLMG iff $F(\mathcal M)$ is a mutex group, that is: for every state reachable by the application of actions, at most one fact $f \in F(\mathcal M)$ is true.\\
    A LMG candidate is a LMG iff every instantiation of its fixed variables leads to a PLMG.
\end{definition}

A mutex group requires solely that \emph{at most one} fact of the group is true in every reachable state.
In many cases, it is also the case that \emph{at least one} fact in the group is true.
We call such mutex groups, LMGs, and PLMGs, \emph{exactly-one} mutex groups, LMGs, and PLMGs.
The LMG inference mechanism based on Lifted Fact Altering Mutex Groups we use \cite{fiser-aaai2020}, determines for every LMG whether it is an exactly-one or an at-most-one LMG.
This naturally also applies to generated PLMGs.

\ifextendedversion
\begin{example}
Consider the transport domain. 
One of the LMGs of this domain is : $\langle \{?p\}, \{?l,?v\}, \{at(?p,?l), in(?p,?v)\} \rangle$. 
From it, we can generate as many PLMGs as there are packages.
For each package $p$, we then obtain a PLMG group $\mathcal M$ with $\mathfrak C(\mathcal M) = \{?l,?v\}$ and $\mathfrak L(\mathcal M) = \{at(p,?l), in(p,?v)\}$.
\label{example:lifted_fam_groups}
\end{example}
\fi

\paragraph{Boolean Satisfiability} Boolean Satisfiability (SAT) is the task of determining whether a propositional formula in conjunctive normal form (CNF) has a truth assignment of its variables that makes the formula true.
A CNF formula consists of a conjunction ($\wedge$) of clauses, where each clause is a disjunction ($\vee$) of literals.
A literal is a propositional variable, or variable for short, $x$ or its negation $\neg x$.
SAT solvers, which are readily available, determine whether a formula is satisfiable, and if so, provide a satisfying truth assignment.
For ease of notation, we will not present SAT formulae in CNF, but all formulae can be translated easily to CNF.

\section{Related Work}

SAT-based planning has been a prominent technique since the work of \citeauthor{kautz-selman-ecai1992}~(\citeyear{kautz-selman-ecai1992}). Most approaches follow a common idea: generate SAT formulas $\phi_\ell$ satisfiable iff a plan of length $\leq \ell$ exists. These are passed to a SAT solver either incrementally (for optimal planning) or via a schedule (e.g., \cite{rintanen-aij2012,rintanen-ipc2011} for satisficing planning). Once a satisfiable $\phi_i$ is found, a plan is extracted.
Several SAT planning systems have since been developed, including SATPLAN04~\cite{kautz-et-al-ipc2006}, Madagascar~\cite{rintanen-aij2012}, Aquaplanning
\footnote{https://github.com/domschrei/aquaplanning}
, and SASE~\cite{huang-et-al-jair2012}, all relying on fully grounded representations. While grounding simplifies reasoning, it poses scalability issues in complex and large domains.

To mitigate this, alternative approaches emerged that avoid full grounding.
Early approaches such as \emph{operator splitting} were introduced by \citeauthor{kautz-et-al-kr1996}~(\citeyear{kautz-et-al-kr1996}) and \citeauthor{Ernst1997AutomaticSAT}~(\citeyear{Ernst1997AutomaticSAT}). \citeauthor{kautz-et-al-kr1996}'s simple splitting assigned each lifted action its own set of argument variables, whereas \citeauthor{Ernst1997AutomaticSAT} proposed an \emph{overloaded} version where all actions shared the same argument variables. \citeauthor{robinson-et-al-icaps2009}~(\citeyear{robinson-et-al-icaps2009}) later extended this idea to support action parallelism by grouping arguments. However, all of these approaches fully grounded predicates and states, as argued for by \citeauthor{kautz-et-al-kr1996}~(\citeyear{kautz-et-al-kr1996}).

LiSAT~\cite{hoeller-behnke-icaps2022} avoids this limitation by using a fully lifted, stateless encoding inspired by plan-space planning. It does not track the state over time, instead, it ensures that each action’s preconditions are achieved by prior actions (or the initial state) and not invalidated by intervening actions. This removes the need to ground facts.

Problematically, LiSAT’s encoding leads to quadratic growth in formula size with respect to plan length $L$, as every precondition must be linked to a potential achiever at any earlier step. This limits LiSAT’s scalability in problems requiring long plans, despite strong performance in the length-optimal setting. LiSAT is also not able to exploit inherent problem structures, like LMGs -- which can cause a substantial overhead in reasoning.

\section{Lifted Representation of Actions}

\citeauthor{villaret2021exploring}~(\citeyear{villaret2021exploring}) introduced a new way of handling action's arguments which LiSAT also uses, i.e., a new way of operator splitting -- \emph{Unified Arguments}.
It is similar to \citeauthor{Ernst1997AutomaticSAT}~(\citeyear{Ernst1997AutomaticSAT})'s overloaded splitting in that it allows different actions to share argument variables.
It however does not perform this sharing simply by argument index.
Instead, it considers the argument's types and shares only arguments of the same type, irrespective of their positions in the action's argument lists.
This is especially useful in domains where the number of objects within types is vastly different.
For each type, the total number of arguments across all actions is computed, and the maximum count is taken as the number of placeholders required for that type.
This process eliminates variability in argument order across actions, as each placeholder is uniquely associated with a type.
By doing so, the formula becomes independent of the specific argument ordering of individual actions.
 

\ifextendedversion
\begin{example}
Consider the actions $drive$($?v - vehicle$, $?l1 - location$, $?l2 - location$), $drop$($?v - vehicle$, $?l - location$, $?p - package$) and $pickup$($?v - vehicle$, $?l - location$, $?p - package$) from the transport domain. 

There are three distinct types involved in this example: vehicle (v), location (l), and package (p). Let $num_{type}(action)$ denote the number of arguments of a specific type that an action takes. The total occurrences of each type in each action are summarized as follows:

\begin{itemize}
    \item $num_{v}(drive)$ = $num_{v}(drop)$ = $num_{v}(pickup)$ = 1
    \item $num_{l}(drive)$ = 2, $num_{l}(drop)$ = $num_{l}(pickup)$ = 1
    \item $num_{p}(drive)$ = 0, $num_{p}(drop)$ = $num_{p}(pickup)$ = 1
\end{itemize}


The resulting Unified Arguments consist of four placeholders: one for vehicles, two for locations, and one for packages. Table~\ref{tab:unified_arguments} illustrates more clearly the alignment of the arguments from all actions with the final UA.

\begin{table}
    \centering
        \begin{tabular}{c|cccl}
             drive&  ?v &  ?l &  ?l &\\
             drop&  ?v &  ?l &  &?p \\
             pick-up&  ?v &  ?l &  &?p \\ \hline \hline
             UA&  vehicle&  location&  location&package\\
        \end{tabular}
    \caption{Unified Arguments}
    \label{tab:unified_arguments}
\end{table}

\label{example:unified_arguments}
\end{example}
\fi

As we will base our own encodings on the concepts of LiSAT including its Unified arguments, we briefly review the relevant sections of LiSAT's formula, that we retain for our work.
%
To encode actions and their arguments, \citeauthor{hoeller-behnke-icaps2022}~(\citeyear{hoeller-behnke-icaps2022}) introduced two types of variables:
\begin{compactitem}
    \item \(a^{t}\): indicates if action $a$ is applied at time step $t$.
    \item \((v=o)^t\): indicates whether an object $o$ is selected by unified argument $v$ or not at time step $t$.
\end{compactitem}
We write $\mathcal{A}^t$ to denote the set of all \(a^{t}\) variables for time step $t$ and $\mathds{UA}_v^t$ to denote the set of all \((v=o)^t\) variables for unified argument $v$ at time $t$.
We write $\mathds{U}$ to denote the set of all unified arguments.
Lastly, let $args$ be a function mapping an action to the set of unified arguments it has as its arguments.
We need to ensure that at most one action, with properly selected arguments is executed at each time step $t$. For this we use the $atMostOne$ macro, which given a set of decision variables generates clauses ensuring at most one is true, yielding the constraints: $atMostOne(\mathcal{A}^{t})$, $atMostOne(\mathds{UA}_v^t)$, and $a^{t} \implies \bigvee_{x \in \mathds{UA}_v^t} x$. Depending on the number of variables, the encoding of the $atMostOne$ constraint varies. For up to 256 variables, a pairwise quadratic encoding is used, whereas a binary counter encoding is applied otherwise~\cite{sinz2005towards}. 
We retain all constraints pertaining to static preconditions described in \citeauthor{hoeller-behnke-icaps2022}~(\citeyear{hoeller-behnke-icaps2022}) and omit them for brevity.

We further retain the constraint $a^t \!\!\!\implies\!\!\! \bigvee_{\forall b \in \mathcal{A}} b^{t-1}$, which reduces symmetries in the formula by enforcing that the time steps with actions are compact at the start of the plan.

\section{Partial Grounding in Propositional Logic}

We present three SAT-based encodings which, through the use of partial grounding, eliminate the quadratic scaling present in LiSAT~\cite{hoeller-behnke-icaps2022}.
The main cause of LiSAT's quadratic formula size was the need to encode causal links, which then was caused by not tracking the state explicitly.
Instead, we will encode the state explicitly, removing the need for encoding causal links and thus ensuring a linear growth behavior of our formula. Our formulas take the overall form $F = F_{\mathcal{I}} \wedge F_{\mathcal{G}} \wedge \bigwedge_{t=0}^\ell \tau(t,t+1)$, where $\tau(t,t+1)$ encodes the transition from time $t$ to $t+1$. The encodings presented throughout the rest of the paper all specify $\tau(t,t+1)$. Since their size is not dependent on the number of time steps, $|\tau(t,t+1)|$ is a constant with respect to $\ell$ and only scales in $|\Pi|$. So $F$ scales linearly in $\ell$.

Our overall approach is to always leave actions fully lifted, while fully or partially (depending on the encoding) grounding the state.
This distinction is reasonable, as the arity of predicates is typically bounded by the arity of the actions they appear in -- which means that predicates and thus the state description, results in a smaller grounding than the actions.
By retaining actions at a lifted level, we preserve LiSAT's compact representation for actions.
Our three encodings differ in the degree in which they ground the state representation, with one fully grounding predicates and the other two partially grounding them using PLMGs, but differing in the way that they encode them.

\subsection{Encoding with Fully Grounded Facts}
We start by giving a simplistic approach that acts as a baseline.
In it, we fully ground all predicates and use them as the state representation.
This encoding is in fact similar to the one proposed by \citeauthor{kautz-et-al-kr1996} and \citeauthor{kautz-selman-aaai1996}~(\citeyear{kautz-et-al-kr1996}), as operationalised and automated by \citeauthor{Ernst1997AutomaticSAT}~(\citeyear{Ernst1997AutomaticSAT}).
The main difference between their encoding and ours are: (1) we use Unified Arguments instead of simple or overloaded splitting, (2) we use LiSAT's~\cite{hoeller-behnke-icaps2022} handling of typing and type constraints, and (3) we use LiSAT's dedicated encoding of static preconditions.
In addition to the variables \(a^{t}\) and \((v=o)^t\) to encode the selected actions, we introduce the following variables to describe the state:
\begin{compactitem}
    \item \(f^{t}\): indicates if fact $f$ is true at time step $t$.
    \item \(c_{a,p(v_1,...,v_n)}^{f,t}\): indicates that the action $a$ makes via its effect $p(v_1,...,v_n)$ the fact $f$ true at time step $t$, with $f$ being the grounding of $p(v_1,...,v_n)$.
\end{compactitem}

For this encoding, we fully ground all predicates to obtain the set $\mathcal{F}$ of all facts.
Further assume we encode the plan for length $L$ and there is a function, $achievers$ which receive a fact as an argument and returns an action and the predicate in its effects (either add or delete) corresponding to the fact.

Firstly, we assert the initial and goal states at time steps $0$ and $L$, respectively, via 
    $\mathcal F_I = \bigwedge_{f \in \mathcal{I}} f^0 \wedge \bigwedge_{f \notin \mathcal{I}} \neg f^0 $ and $\mathcal F_G = \bigwedge_{f \in \mathcal{G}} f^L 
    $.
%
Constraint~(\ref{constraint:precImplication}) forces preconditions of selected actions to hold. Formulae with a parameter $t$ are generated per time step.
Note that $p(o_1,...,o_n)$ in this case is a specific fact.
\begin{align}
    & \forall a \!\in\! \mathcal{A} \forall p(v_1,...,v_n) \!\in\! prec(a) \forall o_1 \!\in\! \mathcal O^{v_1},...,o_n \!\in\! \mathcal O^{v_n}: \nonumber \\ & a^{t} \wedge \bigwedge_{i=1}^{n} (v_i=o_i)^{t} \implies p(o_1,...,o_n)^t \label{constraint:precImplication}
\end{align}
Constraint~(\ref{constraint:effImplication}) ensures that effects of actions take effect at the next time.
The negation ($\neg$) is present for delete effects, but not for add effects.
Constraint~(\ref{constraint:causeVars}) introduces the cause variables that keep track of reasons for a fact to become true (no $\neg$) or false ($\neg$) -- depending on whether the effect is adding or deleting the fact.
\begin{align}
    & \forall a \!\in\! \mathcal{A} \forall (\neg)p(v_1,...,v_n) \!\in\! \mathit{eff}(a) \forall o_1 \!\in\! \mathcal O^{v_1},...,o_n \!\in\! \mathcal O^{v_n}: \nonumber \\ & a^t \wedge \bigwedge_{i=1}^{n} (v_i=o_i)^{t} \implies (\neg)p(o_1,...,o_n)^{t+1}\label{constraint:effImplication}\\
    & c_{a,p(v_1,...,v_n)}^{(\neg)p(o_1,...,o_n),t} \implies a^t \wedge \bigwedge_{i=1}^{n} (v_i=o_i)^t \label{constraint:causeVars}
\end{align}

Finally, Constraints~(\ref{constraint:frameAxiomDel}) and~(\ref{constraint:frameAxiomAdd}) 
specify that a change in a fact's truth is caused by an action's effects.
\begin{align}
    & \forall f \in \mathcal{F}:  f^t \wedge \neg f^{t+1} \implies\hspace{-1cm} \bigvee_{a,p(v_1,...,v_n) \in achievers(\neg f)}\hspace{-1cm} c_{a,p(v_1,...,v_n)}^{\neg f,t+1} \label{constraint:frameAxiomDel}\\
    & \forall f \in \mathcal{F}:\neg f^t \wedge f^{t+1} \implies \hspace{-0.9cm}\bigvee_{a,p(v_1,...,v_n) \in achievers(f)}\hspace{-0.9cm} c_{a,p(v_1,...,v_n)}^{f,t+1} \label{constraint:frameAxiomAdd}
\end{align}

\ifextendedversion
\begin{example}
    Consider a simple planning problem from the transport domain with only 3 objects: v - vehicle, p - package, l - location. In the initial state, $v$ is at $l$ and $p$ is in $v$. In the goal state, $p$ is at $l$. Our encoding of length 1 for this problem would contain the following clauses (the full encoding would have more clauses but we only present the relevant ones for brevity):

    \begin{itemize}
        \item Precondition check:
        \begin{itemize}
            \item $drop^1 \wedge (v_1 = v)^1 \wedge (v_2 = l)^1 \implies at(v,l)^0$
            \item $drop^1 \wedge (v_1 = v)^1 \wedge (v_4 = p)^1 \implies in(p,v)^0$
        \end{itemize}
        \item Effects:
        \begin{itemize}
            \item $drop^1 \wedge (v_1 = v)^1 \wedge (v_4 = p)^1 \implies \neg in(p,v)^1$
            \item $drop^1 \wedge (v_2 = l)^1 \wedge (v_4 = p)^1 \implies at(p,l)^1$
        \end{itemize}
        \item Cause Variables:
        \begin{itemize}
            \item $c_{drop,at(v_2,v_4)}^{at(p,l),1} \implies drop^1 \wedge (v_2 = l)^1 \wedge (v_4 = p)^1$
            \item $c_{drop,at(v_1,v_4)}^{\neg in(p,v),1} \implies drop^1 \wedge (v_1 = v)^1 \wedge (v_4 = p)^1$
        \end{itemize}
        \item Frame Axioms:
        \begin{itemize}
            \item $in(p,v)^0 \wedge \neg in(p,v)^1 \implies c_{drop,at(v_1,v_4)}^{\neg in(p,v),1}$
            \item $\neg at(p,l)^0 \wedge at(p,l)^1 \implies c_{drop,at(v_2,v_4)}^{at(p,l),1}$
        \end{itemize}
    \end{itemize}
    
\end{example}
\fi

\subsubsection{Predicate Pruning}
As an important optimisation to all our encodings, we remove facts $f$ from the list of all facts $\mathcal{F}$ whose truth value does not matter for any plan.
Intuitively, a fact can only influence a plan if it appears either in the goal or as a precondition.
Facts that do not occur in either place cannot affect the success or failure of any action execution or the satisfaction of the goal.
They can therefore be safely discarded.
Such irrelevant facts are commonly pruned in grounded planning (see e.g.~\cite[Def.\ 8]{muise-et-al-icaps2012} for grounded probabilistic planning).
Due to lifting, we cannot analyze each ground fact individually.
Instead, we remove entire predicates that do not appear in any precondition.
This might over-delete facts that appear in no precondition, but appear only in the goal.
As an exception, goal facts are added as individual facts to be considered when encoding.
This way ensures that the goal conditions are checked correctly.
We refer to this as \emph{predicate pruning}.



\subsection{Encoding with Partially Grounded Facts}
In this section, we present the novel contributions of this work. We introduce a new family of SAT encodings that leverage partially lifted mutex groups (PLMGs) to partially ground the state description while keeping the action encoding fully lifted. This approach exploits the inherent structure of planning problems to improve the scalability of the encoding. Unlike previous encodings of lifted planning into SAT, whose formula size scales quadratically with plan length, our encoding's formula size scales linearly, therefore addressing a key limitation of existing methods.

In practice, the fully grounded encoding leads to prohibitively large formulae.
This can e.g. be caused by a high number of objects (more than $10^3$) or by predicates of high arity (greater than 2) that cause a lot of facts in grounding.
One way to counteract this, is to exploit the planning problem's inherent internal structure.
For this, we leverage PLMGs -- they will allow us to only ``partially'' ground the predicates enabling a smaller representation.

Let $\mathcal M$ be a PLMG.
From it, we know that in every state at most one of the ground instances (i.e.\ regular facts) of its literals can be true.
Instead of generating all these literals, we use the counted variables of the PLMG to implicitly denote which of the facts is currently true.
To do so, we need to (1) identify from which atom in $\mathfrak L(\mathcal M)$ that fact was grounded and (2) what the value of the counted variables in it is.
This can allow for a more compact representation.


In practice the LMG inference by \citeauthor{fiser-aaai2020}~(\citeyear{fiser-aaai2020}), generates more LMGs than necessary and notably overlapping LMGs.
We use~\citeauthor{helmert-aij2009}'s~(\citeyear{helmert-aij2009}) greedy PLMG selection, which has been the standard for $\geq$15 years. Non-greedy alternatives typically yield negligible or even negative impact on performance.
First, if the grounding of a LMG contains all facts of a predicate $p$, we select all PLMGs grounded from it and remove $p$ from consideration.
We repeat this process until no further such LMGs can be found.
We then fully ground the remaining predicates (typically a small subset of all predicates, or none at all) and LMGs into PLMGs.
We select PLMGs greedily, starting with the ones that cover the most still uncovered facts, until no PLMGs that would cover new facts remain.
As a result, we obtain a set $\mathfrak P$ of PLMGs that we use for encoding.
After this procedure, there might still be facts $\mathfrak U$ not covered by any of the PLMGs.
For simplicity, we assume that no PLMG contains two literals with the same predicate, i.e., that they can be identified by their predicate, which is true for all LMG group generated by \citeauthor{fiser-aaai2020}~(\citeyear{fiser-aaai2020}).


For the encoding, we 
retain the variables $f^t$ and the cause variables only for the uncovered facts in $\mathfrak U$. For them, we also retain the encoding previously described.
We remove the facts pertaining to all other facts and encode them using PLMGs instead.
For this, we introduce new variables:
\begin{compactitem}
    \item \((c^{\mathcal{M}} = o)^t\): indicating whether the counted variable $c$ of the PLMG $\mathcal M$ has the value $o$ at time $t$.
    \item \((p^\mathcal M)^t\): indicating that the literal of predicate $p$ of PLMG $\mathcal M$ is true at time $t$.
    \item \((v_j \!\in\! D(c^{\mathcal{M}}))^t\): indicating whether the object assigned to the unified argument $v_j$ is of the same type of the counted variable $c$ of PLMG $\mathcal M$ at time $t$.
    \item \((c^{\mathcal{M}} \!\equiv\! v_j)^t\): indicating if counted variable $c$ of PLMG $\mathcal M$ has the same value as the unified argument $v_j$ at time $t$.
    \item $(c^{\mathcal{M}} \leftarrow a_{p(v_1,\dots,v_n)} )^t$: indicating that action $a$ has set the value of counted variable $c^{\mathcal{M}}$ via its effect $p(v_1,\dots,v_n)$.
    \item $(p^{\mathcal{M}} \!\!\!\leftarrow\!\! a_{p(v_1,\dots,v_n)})^t$: indicating action $a$ has set the predicate of PLMG $\mathcal M$ to $p$ via its effect $p(v_1,\dots,v_n)$.
    \item $(c^{\mathcal{M}} \not \equiv c^{\mathcal{M}})^t$: indicating that the value of the counted variable $c$ of PLMG $\mathcal M$ is different at times $t-1$ and $t$.
\end{compactitem}

For notational purposes, we define the function $\textbf{O}$, which takes a literal of a PLMG as an argument.
It returns a set of pairs $(o,j)$ for each object $o$ at the $j$th argument position of that literal.
We introduce $\textbf{V}$ to return a set of pairs $(v,j)$ for each variable $v$ at the $j$th argument position of that literal.

First, we assert per PLMG that every counted variable has exactly one value and that exactly one of the literals is selected.
This constraint is only correct for exactly-one PLMGs.
For PLMGs that are not exactly-one, we add a new literal "none" to that PLMG that is true iff none of the other values in the PLMG are true.
With it, we can force also for these PLMGs that exactly one literal of the PLMG is true.
%
\begin{align}
&\forall \mathcal{M} \in \mathfrak P \ \ \forall c \in \mathfrak C(\mathcal{M}) : \nonumber\\
&atMostOne(\{(c^{\mathcal M} \!=\! o)^t \!\mid\! o \!\in\! \mathcal O^c\}) \!\land\!\! \bigvee_{o \in \mathcal O^c} \!\!(c^{\mathcal M} = o)^t\\
&\forall \mathcal{M} \in \mathfrak P:\nonumber\\
&atMostOne(\{(p_i^{\mathcal M})^t \mid p(x_{1},\dots,x_{m}) \in \mathfrak L(\mathcal{M})\})\\
&\bigvee_{p(x_{1},\dots,x_{m}) \in \mathfrak L(\mathcal{M})} (p_i^{\mathcal M})^t \label{eq:atleastoneplmg}
\end{align}

As the second step, we need to correctly assert the initial state.
For this, we iterate over all facts in the initial state and PLMG and check whether it contains a matching literal, if so, we assert it.
Matching here means that if both have objects at a given argument position, they are the same, and otherwise the object in the initial state's fact is part of the type of the PLMG's literal's variable.
We also generate symmetric clauses for the goal, which we here omit for brevity.
\begin{align}
    & \forall p(o_1,...,o_{m}) \!\in\! \mathcal{I} \ \ \forall \mathcal{M} \in \mathfrak P \ \ \forall p(x_{1},\dots,x_{m}) \in \mathfrak L(\mathcal{M}): \nonumber \\
    &\text{if } \forall (o,j) \in \textbf{O}(p(x_{1},...,x_{m_i}))\!:\!o = o_j \nonumber\\
    &\text{and } \forall (c,j) \in \textbf{V}(p(x_{1},...,x_{m_i})) : o_j \!\in\! \mathcal O^c \text{ then} \nonumber \\
    & (p^{\mathcal M})^0 \wedge \bigwedge_{(c,j) \in \textbf{V}(p(x_{1},\dots,x_{m}))} (c^{\mathcal M} = o_j)^0 \label{constraint:pgfInit}
\end{align}

Thirdly, we have to assert that preconditions of actions are actually met.
This means that for every PLMG containing the precondition with its chosen arguments as a grounding, this grounding must be true in the previous time step in one of the PLMG.
Here, we do not access the values of the PLMG's counted variables directly, but assert the intermediate \((c^{\mathcal{\mathcal M}} \equiv v_j)^t\) variables to be true, stating that a specific counted variable must equal a specific unified argument.

\begin{align}
    & \forall a \!\in\! \mathcal{A} \forall p(v_1,\dots,v_m) \!\in\! prec(a) \nonumber\\
    & \forall \mathcal{M} \in \mathfrak P \ \ \forall p(x_{1},\dots,x_{m}) \in \mathfrak L(\mathcal{M}): \nonumber\\
    &\text{if } \forall (o,j) \!\in\! \textbf{O}(p(x_{1},...,x_{m})): o \!\in\! \mathcal O^{v_j} \text { then}\nonumber \\
    & a^{t} \wedge \!\!\!\!\!\!\!\!\bigwedge_{(o,j) \in \textbf{O}(p(x_{1},\dots,x_{m}))} \!\!\!\!\!\!\!\!\!\!(v_j = o)^{t} \wedge \!\!\!\!\!\!\!\!\bigwedge_{(c,j) \in \textbf{V}(p(x_{1},\dots,x_{m}))} \!\!\!\!\!\!\!\!\!\!\!\!\!\!(v_j \!\in\! D(c))^{t} \nonumber\\
    & \hspace{2cm} {} \implies (p^{\mathcal{M}})^t \wedge \!\!\!\!\!\!\!\!\bigwedge_{(c,j) \in \textbf{V}(p(x_{1},\dots,x_{m}))} \!\!\!\!\!\!\!\!\!\!\!\!\!\!(c^{\mathcal{M}} \equiv v_j)^t \label{constraint:pgfPrec}
\end{align}
Similarly, we have to assert that the effects of actions take place at the same time step and thus set the value of the counted variables of PLMGs to specific values.
Note that an effect can either be positive (adding) or negative (deleting). In the latter case, we need to negate the implication's right-hand-side to ensure that the PLMG does not take the deleted value at the next time step.
%


\begin{align}
    & \forall a \!\in\! \mathcal{A} \forall (\neg) p(v_1,\dots,v_n) \!\in\! \mathit{eff}(a) \nonumber\\
    & \forall \mathcal{M} \in \mathfrak P \ \ \forall p(x_{1},\dots,x_{m}) \in \mathfrak L(\mathcal{M}): \nonumber\\
    & \text{if } \forall (o,j) \!\in\! \textbf{O}(p(x_{1},...,x_{m})): o \!\in\! \mathcal O^{v_j} \text{ then} \nonumber \\ & a^t \wedge \!\!\!\!\!\!\!\!\bigwedge_{(o,j) \in \textbf{O}(p(x_{1},\dots,x_{m}))} \!\!\!\!\!\!\!\!\!\!(v_j = o)^t \wedge \!\!\!\!\!\!\!\!\bigwedge_{(c,j) \in \textbf{V}(p(x_{1},\dots,x_{m}))} \!\!\!\!\!\!\!\!\!\!\!\!\!\!(v_j \!\in\! D(c))^{t} \nonumber\\ & \hspace{0.5cm} {} \implies (\neg) \left[(p^{\mathcal{M}})^{t+1} \wedge \!\!\!\!\!\!\!\!\bigwedge_{(c,j) \in \textbf{V}(p(x_{1},\dots,x_{m}))} \!\!\!\!\!\!\!\!\!\!\!\!\!\!(c^{\mathcal{M}} \equiv v_j)^{t+1} \right]\label{constraint:pgfEff}
\end{align}

It is possible that for a PLMG an action with a delete effect on it but no add effect is executed.
In this case, the PLMG becomes inactive, i.e., all of its ground facts are false.
In our encoding the literal "none" then becomes true -- forced by Eq.~\ref{eq:atleastoneplmg}.
The frame axioms introduced next ensure that no other literal in the PLMG can be selected.
For the LMGs generated via Lifted Fact Altering Mutex Groups~\cite{fiser-aaai2020}, it is further the case that once a PLMG has become inactive it can never become active again, i.e., once the "none" literal is true it will stay true~\cite[Thm. 6]{fiser-aaai2020}.
We enforce this using the formula $(\text{none}^{\mathcal{M}})^{t} \implies (\text{none}^{\mathcal{M}})^{t+1}$.
If one were to use LMGs generated by a different method, this constraint must not be added.
The remaining encoding naturally allows the "none" to become false again and sets the correct ground fact of the PLMG to true.

As the fourth step, we create the frame axioms, ensuring that the state does not change from one time to the next, except if caused by an action's effects.
For representing the state using PLMGs, this means that the value of a counted variable can only change if there is an action that has set it to a specific value (i.e.\ object).
There are two possible ways of encoding this: either we encode that for a variable to attain a value an action must have set it to that value, or we encode that if the variable’s value changed (to any value), an action must have set it.
The first encoding is quite cumbersome and large, as it needs to create clauses for every possible value that the counted variable could have.
Further, the encoding of effects already uses variables of the type $(c^{\mathcal{M}} \equiv v_j)^{t}$ which if true, mean the variable $c$ was set to a specific value, namely the value of the unified argument $v_j$.
We first encode the causes that can change the variable's value as expressed by the $(c^{\mathcal{M}} \leftarrow a_{p(v_1,\dots,v_n)} )^t$ and $(p^{\mathcal{M}} \leftarrow a_{p(v_1,\dots,v_n)})^t$ variables, which we call \emph{cause variables}.
\begin{align}
    & \forall a \!\in\! \mathcal{A} \forall p(v_1,\dots,v_n) \!\in\! \mathit{eff}(a) \nonumber\\
    & \forall \mathcal{M} \in \mathfrak P \ \ \forall p(x_{1},\dots,x_{m}) \in \mathfrak L(\mathcal{M}): \nonumber\\
    & \text{if } \forall (o,j) \!\in\! \textbf{O}(p(x_{1},...,x_{m})): o \!\in\! \mathcal O^{v_j} \text{ then} \nonumber \\
    & a^t \wedge \!\!\!\!\!\!\!\!\bigwedge_{(o,j) \in \textbf{O}(p(x_{1},\dots,x_{m}))} \!\!\!\!\!\!\!\!\!\!(v_j = o)^t \wedge \!\!\!\!\!\!\!\!\bigwedge_{(c,j) \in \textbf{V}(p(x_{1},\dots,x_{m}))} \!\!\!\!\!\!\!\!\!\!\!\!\!\!(v_j \!\in\! D(c))^{t} \nonumber\\
    & \hspace{2cm} {} \implies (p^{\mathcal{M}} \leftarrow a_{p(v_1,\dots,v_n)} )^{t+1} \wedge{} \nonumber\\
    &\hspace{2.2cm} {}\bigwedge_{(c,j) \in \textbf{V}(p(x_{1},\dots,x_{m}))} \!\!\!\!\!\!\!\!\!\!\!\!\!\!(c^{\mathcal{M}} \leftarrow a_{p(v_1,\dots,v_n)} )^{t+1}\\
    & (p^{\mathcal{M}} \leftarrow a_{p(v_1,\dots,v_n)})^{t+1} \lor (c^{\mathcal{M}} \leftarrow a_{p(v_1,\dots,v_n)} )^{t+1} \implies \nonumber\\
    &a^t \wedge \!\!\!\!\!\!\!\!\bigwedge_{(o,j) \in \textbf{O}(p(x_{1},\dots,x_{m}))} \!\!\!\!\!\!\!\!\!\!(v_j = o)^t \wedge \!\!\!\!\!\!\!\!\bigwedge_{(c,j) \in \textbf{V}(p(x_{1},\dots,x_{m}))} \!\!\!\!\!\!\!\!\!\!\!\!\!\!(v_j \!\in\! D(c))^{t}
\end{align}
We have to connect the causes for a counted variable to be set to the change of the value of that variable.
Let $\mathds{C}(c^{\mathcal M}, t)$ be the set of all cause variables for the counted variable $c$.
Any change in literal, except to "none", has to be caused.
\begin{align}
    & \forall \mathcal{M} \in \mathfrak P \ \  \nonumber\\
&\forall c \in \mathfrak C(\mathcal{M}): (c^{\mathcal M} \not \equiv c^{\mathcal M})^{t} \implies \bigvee_{x \in \mathds{C}(c^{\mathcal M},t)} x\\
    &\forall p(x_{1},...,x_{m}) \!\!\in\!\! \mathfrak L(\mathcal{M})\!\!:\!\! \neg(p^{\mathcal M})^{t} \!\land\! (p^{\mathcal M})^{t+1} \!\!\!\implies\!\!\! \hspace{-0.5cm} \bigvee_{x \in \mathds{C}(p^{\mathcal M},t)}\hspace{-0.5cm} x
\end{align}

As the fifth and last step, we need to assert the semantics of the intermediate helper variables \((v_j \!\in\! D(c^{\mathcal{M}}))^t\), \((c^{\mathcal{M}} \equiv v_j)^t\) and $(c^{\mathcal{M}} \not \equiv c^{\mathcal{M}})^t$ that we used throughout the encoding, expressing that the value of a unified argument is within the type of a counted variable, that a counted variable has the same value of a unified argument, and that the value of a counted variable changed, respectively.
\begin{align}
    & \forall \mathcal{M} \in \mathfrak P \ \ \forall c \in \mathfrak C(\mathcal{M}) \ \ \forall v \in \mathds{U}: \nonumber\\
    & \forall o \in \mathcal O^v \setminus \mathcal O^c: (v = o)^t \implies \neg (v \!\in\! D(c^{\mathcal M}))^t \label{constraint:outside_1}\\
        &\forall o \in \mathcal O^v \cap \mathcal O^c:(v = o)^t \implies (v \!\in\! D(c^{\mathcal M}))^t\\
    & \qquad{}\land\!(c^{\mathcal M} = o)^t \!\wedge\! (v = o)^t \!\!\!\implies\!\!\! (c^{\mathcal M} \!\equiv\! v)^t \\
    & \qquad{}\land \!(c^{\mathcal M} \!\equiv\! v)^t \!\implies\! \left[(c^{\mathcal M} = o)^t \!\Leftrightarrow\! (v = o)^t\right] \\
    %
    %
    & \forall o \in \mathcal O^c \setminus \mathcal O^v: (c^{\mathcal M} = o)^t \implies \neg (c^{\mathcal M} \equiv v)^t \label{constraint:outside_2}\\
    & \forall \mathcal{M} \in \mathfrak P \ \ \forall c \in \mathfrak C(\mathcal{M}) \forall o \in \mathcal O^c: \nonumber\\
    &(c^{\mathcal M} = o)^{t-1} \land \neg (c^{\mathcal M} = o)^t  \implies (c^{\mathcal{M}} \not \equiv c^{\mathcal{M}})^t \label{form:cchanges}
    %
    %
    %
    %
    %
\end{align}
It is sufficient to encode only the ``falling edge'' in equation \ref{form:cchanges}, as there is always exactly one $(c^{\mathcal M} = o)^t$ that is true.

\begin{table*}
    \setlength{\tabcolsep}{1mm}
    \centering
    \footnotesize
    \begin{tabular}{l|r@{\hspace{0.5mm}}r|r@{\hspace{0.5mm}}r|r@{\hspace{0.5mm}}r|r|r@{\hspace{0.5mm}}r|r|r@{\hspace{0.5mm}}r|r@{\hspace{0.5mm}}r||r|r|r|r|r@{\hspace{0.5mm}}r|r@{\hspace{0.5mm}}r} 
         &  \multicolumn{2}{|c|}{\rotatebox{90}{Fully}} & \multicolumn{2}{|c|}{\rotatebox{90}{Partially}} & \multicolumn{2}{|c|}{\rotatebox{90}{Binary}} & \rotatebox{90}{LiSAT}& \multicolumn{2}{|c|}{\rotatebox{90}{PWL A\textsuperscript{*}}} & \rotatebox{90}{\begin{minipage}{1.2cm}CPDDL\\A\textsuperscript{*}\end{minipage}}& \multicolumn{2}{|c|}{\rotatebox{90}{MpC}} & \multicolumn{2}{|c||}{\rotatebox{90}{\begin{minipage}{1.2cm}FD A\textsuperscript{*}\\ lmc\end{minipage}}} & \rotatebox{90}{Binary}& \rotatebox{90}{LiSAT}& \rotatebox{90}{\begin{minipage}{1.2cm}PWL\\alt-bfws1\end{minipage}}& \rotatebox{90}{\begin{minipage}{1.2cm}CPDDL\\gbfs\end{minipage}}& \multicolumn{2}{|c|}{\rotatebox{90}{MpC}} & \multicolumn{2}{|c}{\rotatebox{90}{\begin{minipage}{1.2cm}FD\\LAMA\end{minipage}}}\\         
         \cline{2-23}&  -&  PP&  -&  PP&  -&  PP&- & $h^{1}$& lmc& lmc & inv.& & -&AS &  PP&- & hff&  hff& inv.&  & -&AS\\ \hline 
         Blocks (40)&21&21&15&15&\textbf{40}&\textbf{40}&\textbf{40} & 0& 2& 34& 4& 4&12 &12  &\textbf{40}&\textbf{40} & 22& 3& 4& 0&12 &12\\  
         Childs (144)&48&48&\textbf{96}&\textbf{96}&\textbf{96}&\textbf{96}&83& 2& 1& 6& 1& 0&8 &6 &\textbf{144}&\textbf{144}& 104& 47& 66& 66&110&138\\
         GED (312)&67&67&64&64&64&64&\textbf{72}& 32& 0& 36& 56& 62&40 &50&77&110& 286& 88& 183& 77&\textbf{312}&\textbf{312}\\ 
         Logistics (40)&26&26&\textbf{40}&\textbf{40}&\textbf{40}&\textbf{40}&32& 5& 0& 31& 0& 0&16 &13&\textbf{40}&\textbf{40}& \textbf{40}& 13& 0& 0&24&16\\
         OS (56)&54&54&54&54&54&54&\textbf{56} & 44& 24& 32& 0& 0&18 &44&51&\textbf{55}& 50& 31& 0& 0&18 &46\\ 
         Pipes (50)&35&35&\textbf{36}&\textbf{36}&32&34&21 & 9& 9& 15& 7& 6&10 &10&43&25& \textbf{48}& 20& 10& 9&18&37\\
         Rover (40)&5&5&\textbf{16}&\textbf{16}&12&12&4 & 2& 0& 3& 0& 0&5 &5&12&4 & \textbf{40}& \textbf{40}& 0& 0&12&12\\ 
         Visitall (180)&29&34&57&110&56&\textbf{115}&107& 73& 30& 62& 18& 28&71 &68&\textbf{178}&176& 147& 121& 18& 56&90&90\\
         Labyrinth (20)&8&8&8&8&8&8&\textbf{12} & 1& 0& 1& 0& 2&1 &8&8&12& 5& 0& 3& 1&\textbf{15}&3\\ \hline \hline
         Cov. (882)& 293& 298& 386& 439& 402& \textbf{463}&427& 168& 66& 220& 86& 102&181 &216& 593&606& \textbf{742}& 363& 284& 209&611&666\\ \hline
         Score (9)& 4.07& 4.10& 5.05& 5.34& 5.49& \textbf{5.86}&5.32& 1.72& 0.83& 3.12& 0.53& 0.68&1.97 &2.72 & 6.71&6.51& \textbf{7.11}& 3.63& 1.59& 1.25&4.90&5.17\\ 
    \end{tabular}
    \caption{Coverage results for length-optimal (left) and satificing (right) planning. Maxima for each group are in bold.}
    \label{tab:results}
\end{table*}

\ifextendedversion
\begin{example}
    Consider the same planning problem as in the previous example. Assume there are the following PLMGs: $\mathcal M: \langle \{p\}, \{?v,?l\}, \{in(p, ?v), at(p,?l)\}\rangle$ and $m: \langle \{v\}, \{?l\}, \{at(v,?l)\}\rangle$. 
    \begin{itemize}
        \item Precondition check:
        \begin{itemize}
            \item $drop^1 \wedge (v_4 = p)^1 \wedge (v_1 \in D(v\_cnt))^1 \implies (in^{\mathcal{M}})^0 \wedge (v\_cnt==v_1)^0$
            \item $drop^1 \wedge (v_1 = v)^1 \wedge (v_2 \in D(l\_cnt))^1 \implies (at^{m})^0 \wedge (l\_cnt==v_2)^0$
        \end{itemize}
        \item Effects: Same logic as preconditions, just a different time step at the right side of the implication.
        \item Frame Axioms:
        \begin{itemize}
            \item  $drop^1 \wedge (v_4 = p)^1 \wedge (v_2 \in D(l\_cnt))^1 \implies (at^{\mathcal{M}} \leftarrow drop)^1 \wedge (l\_cnt \leftarrow drop)^1$
            \item $(at^{\mathcal{M}} \leftarrow drop)^1 \vee (l\_cnt \leftarrow drop)^1 \implies drop^1 \wedge (v_4 = p)^1 \wedge (v_1 \in D(cnt))^1$
        \end{itemize}
        \item Connecting Cause Variables:
        \begin{itemize}
            \item $({\ensuremath{\mathit{diff}}\xspace}\_l\_cnt)^1 \implies (l\_cnt \leftarrow drop)^1$
            \item $\neg(at^{\mathcal{M}})^0 \wedge (at^{\mathcal{M}})^1 \implies (at^{\mathcal{M}} \leftarrow drop)^1$
        \end{itemize}
        \item Assert intermediate variables:
        \begin{itemize}
            \item $(v_2 = l)^1 \implies (v_2 \in D(l\_cnt))^1$
            \item $(l\_cnt==v_2)^1 \wedge (l\_cnt=l)^1 \implies (v_2 = l)^1$
            \item $(l\_cnt==v_2)^1 \wedge (v_2 = l)^1 \implies (l\_cnt=l)^1$
            \item $(l\_cnt=l)^1 \wedge (v_2 = l)^1 \implies (l\_cnt==v_2)^1$
            \item $(l\_cnt=l)^1 \wedge \neg(l\_cnt=l)^1 \implies ({\ensuremath{\mathit{diff}}\xspace}\_l\_cnt)^1$
        \end{itemize}
    \end{itemize}

Although it can not be demonstrated in a simple example such as this one, constraints~(\ref{constraint:outside_1}) and~(\ref{constraint:outside_2}) as well as variables $(v_j \!\in\! D(c^{\mathcal{M}}))^t$ are fundamental for the encoding in cases where the type of a counted variable in a PLMG is a subtype of the corresponding $\mathds{UA}$, ensuring that impossible implications are not forced.
\end{example}
\fi

\subsection{Binary Encoding of Objects}
In the encoding with partially grounded facts, the information density contained in the variables $(c^{\mathcal M} = o)^t$ is fairly small since  at most one of the variables for the admissible objects $o$ is true.
As such, this encoding requires linearly many variables -- which can be a hindrance, especially in planning problems with many objects.
Given a set of objects $O$, it is possible to encode the information contained in the $(c^{\mathcal M}=o)^t$ variables with only $\lceil \log_2(|O|)\rceil$ variables via a bit-representation. 
This increases the information density of valuations of the SAT formula we generate, but simultaneously also decreases the overall formula size in terms of number of clauses.
This observation was, as far as we know, first made by \citeauthor{Ernst1997AutomaticSAT}~(\citeyear{Ernst1997AutomaticSAT}).
They applied it only to the set of grounded actions.

To simplify formula generation, we initially assume that all variables can have any object as its value.
Further, we assume the set of objects $\mathcal O$ is an ordered set and the objects are numbered $o_0$ to $o_{|\mathcal O| - 1}$.
To denote a specific object, $B \!=\! \lceil \log_2(|\mathcal O|)\rceil$ variables are needed.
\gregor{
We introduce two new types of variables replacing $(v = o)^t$ and $(c^{\mathcal M}\!=\!o)^t$: } 
\begin{compactitem}
    \item $v^t_b$: indicates that the $b$th bit of the index of the object assigned to unified argument $v$ at time $t$ is true.
    \item $(c^{\mathcal M})^t_b$: indicates that the $b$th bit of the index of the object assigned to variable $c^{\mathcal M}$ at time $t$ is true.
\end{compactitem}
For notation, we define the function $\beta(x,b)$ to be true if the $b$th bit of $x$ is 1 and false otherwise.
Similarly, we write $(\neg)^p$ where $p$ is a truth value to indicate that a negation should be present if p is true, and that it should not be present if $p$ is false.
We write $\mathcal B$ for the set $\{1, \dots, B\}$.

We start by connecting the values of the variables representing unified arguments $(v=o)^t$ to the $v^t_i$ variables.
\begin{align} \label {eq:def_bin_ua}
\forall v \forall o_i \in \mathcal O\forall b \in \mathcal B:
    (v=o_i)^t \implies (\neg)^{\neg \beta(i,b)} v^t_b
\end{align}
It would be possible to entirely remove the one-hot representation of the variables $(v_j = o)^t$ for the unified arguments and to remove the clauses in Eq.~\ref{eq:def_bin_ua}.
The reduction from doing so is significantly less noticeable than from replacing the  $(c^{\mathcal M}=o)^t$ variables: the unified arguments only account for a fraction of the overall encoded variables.
On the other hand, encoding constraints on the unified arguments on action level (static preconditions, inequalities) becomes more complicated with a binary-only representation -- which prompted us to retain the one-hot encoding here.

Secondly, we have to encode for each variable $c^\mathcal M$ that the object identified by bitmask in the $(c^{\mathcal M})^t_i$ variables is of the type of that variable.
Let this type be $t(c)$.
After preprocessing, we know that the objects of type $t(c)$ form a subsequence of the sequence of all objects, that is there are upper $u_{t(c)}$ and lower $\ell_{t(c)}$ bounds so that the type $t(c)$ comprises exactly the objects $\{o_i \mid \ell_{t(c)} \leq i \leq u_{t(c)}\}$.
To implement these restrictions, we add the following clauses.
Note that bit 0 is the least significant bit of the representation.
\begin{align}
    &\forall \mathcal{M} \in \mathfrak P \forall c \in \mathfrak C(\mathcal M)
    \forall b \in \mathcal B \nonumber\\
    &\text{ if } \beta(\ell_{t(c)},b):
        \hspace{-0.6cm}\bigwedge_{b' \in \{b+1, \dots B\}}\hspace{-0.5cm} (\neg)^{\neg \beta(\ell_{t(c)},b')} (c^{\mathcal M})_{b'}^t
       \!\!\implies\!\! (c^{\mathcal M})_b^t\\
   &\text{ if } \neg \beta(u_{t(c)},b)\!:\!
        \hspace{-0.7cm}\bigwedge_{b' \in \{b+1, \dots B\}}\hspace{-0.7cm} (\neg)^{\neg \beta(u_{t(c)},b')} (c^{\mathcal M})_{b'}^t
       \!\!\!\implies\!\! \neg (c^{\mathcal M})_b^t
\end{align}
Thirdly, we have to replace the clauses defining the semantics of $(c^{\mathcal{M}} \equiv v_j)^t$. \gregor{We introduce a new type of variable to express bit-wise equality:}

\begin{compactitem}
    \item \gregor{$(c^{\mathcal M} \equiv v_j)^t_b$: indicates that the  $b$th bit of the indices of the objects assigned to $c^{\mathcal M}$ and $v$ at time $t$ is the same.}
\end{compactitem}
\gregor{If the $b$th bit of  $c^{\mathcal M}$ and $v$ are equal, $(c^{\mathcal M} \equiv v_j)^t_b$ is true and if all $(c^{\mathcal M} \equiv v_j)^t_b$ are true, $c^{\mathcal M}$ and $v$ are equal.}
\begin{align}
    &\forall \mathcal{M} \in \mathfrak P \forall c \in \mathfrak C(\mathcal M) \forall v_j \in \mathds{U}\ \ \forall b \in \mathcal B:\nonumber\\
    & v^t_b \land (c^{\mathcal M})^t_b \implies (c^{\mathcal M} \equiv v_j)^t_b\\
    & \neg v^t_b \land \neg (c^{\mathcal M})^t_b \Rightarrow (c^{\mathcal M} \equiv v_j)^t_b\\
    & \left(\bigwedge_{b \in \mathcal B} (c^{\mathcal M} \equiv v_j)^t_b \right) \Rightarrow (c^{\mathcal M} \equiv v_j)^t
\end{align}
\gregor{Conversely, if $(c^{\mathcal M} \equiv v_j)^t$ is true, all bits of $c^{\mathcal M}$ and $v$ must be equal.}
\begin{align}
    &(c^{\mathcal M})^t_b \land (c^{\mathcal M} \equiv v_j)^t \implies v^t_b \\
    & v^t_b \land (c^{\mathcal M} \equiv v_j)^t \implies (c^{\mathcal M})^t_b
\end{align}

Lastly, we have to replace the clauses defining the semantics of the $(c^{\mathcal{M}} \not \equiv c^{\mathcal{M}})^t$ variables, encoding that the value of $c^{\mathcal{M}}$ changed from time $t$ to time $t+1$.
We need to introduce auxiliary variable for every bit:
\begin{compactitem}
    \item $(c^{\mathcal{M}} \not \equiv c^{\mathcal{M}})^t_i$: indicates that the $i$th bit of value of variable $c^{\mathcal{M}}$ is different between times $t$ and $t+1$.
\end{compactitem}
We then add the following clauses $\forall \mathcal M \in \mathfrak P \forall c \in \mathfrak C(\mathcal M)$:
\begin{align}
    &(c^{\mathcal{M}} \not \equiv c^{\mathcal{M}})^t \implies \bigvee_{b \in [1, \dots, B]} (c^{\mathcal{M}} \not \equiv c^{\mathcal{M}})^t_b\\
    &\forall b \in \mathcal B: (c^{\mathcal{M}} \not \equiv c^{\mathcal{M}})^t_b \implies (c^{\mathcal{M}} \not \equiv c^{\mathcal{M}})^t \\
    &\forall b \in \mathcal B: (c^{\mathcal{M}})^t_b \land \neg (c^{\mathcal{M}})^{t+1}_b \implies (c^{\mathcal{M}} \not \equiv c^{\mathcal{M}})^t_b\\
    &\forall b \in \mathcal B: \neg (c^{\mathcal{M}})^t_b \land (c^{\mathcal{M}})^{t+1}_b \implies (c^{\mathcal{M}} \not \equiv c^{\mathcal{M}})^t_b
\end{align}

\section{Experimental Evaluation}
\begin{figure*}[t] 
    \centering
    \begin{subfigure}[b]{0.32\textwidth}
        \includegraphics[width=\linewidth, trim={0 0 0 120pt}, clip]{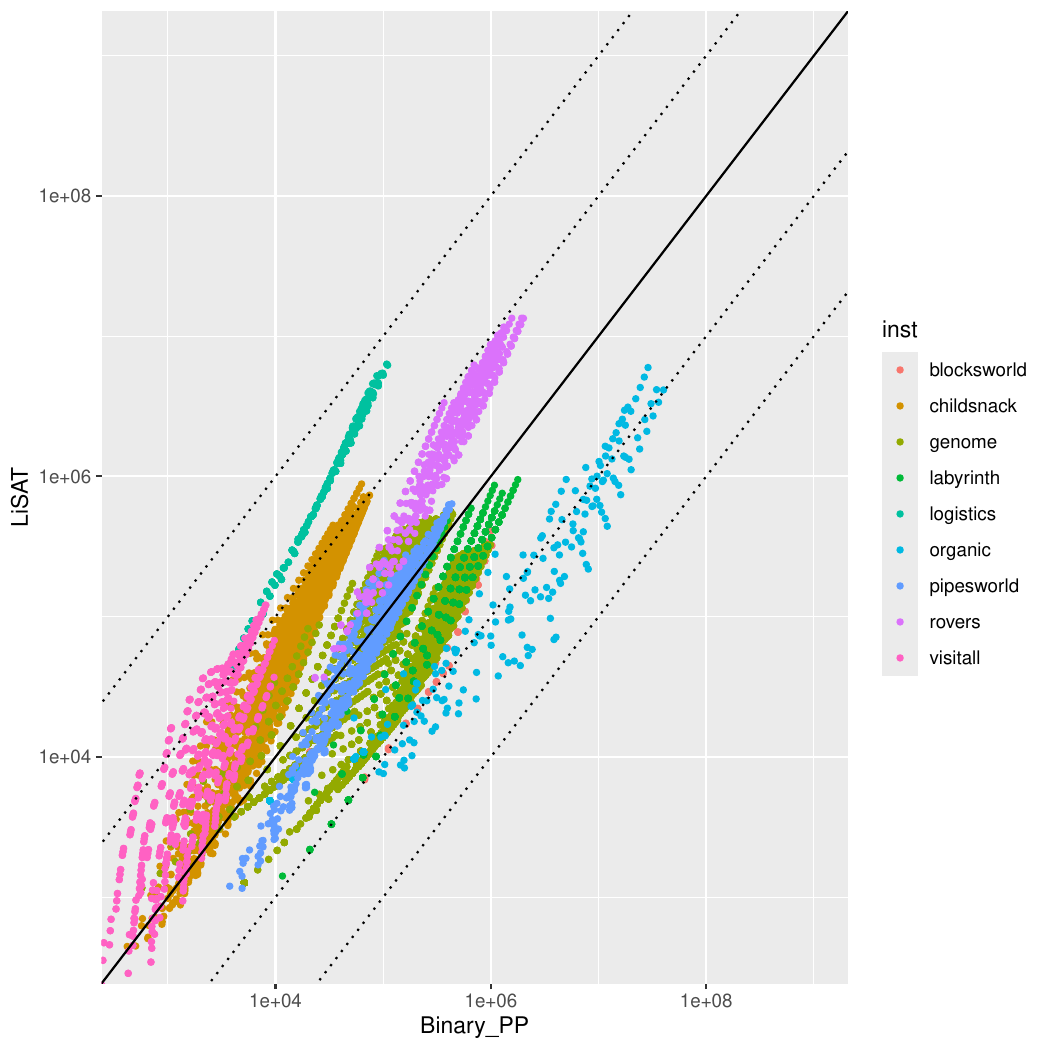}
        \caption{Number of variables (log-log)}
    \end{subfigure}
    \hfill
    \begin{subfigure}[b]{0.32\textwidth}
        \includegraphics[width=\linewidth, trim={0 0 0 120pt}, clip]{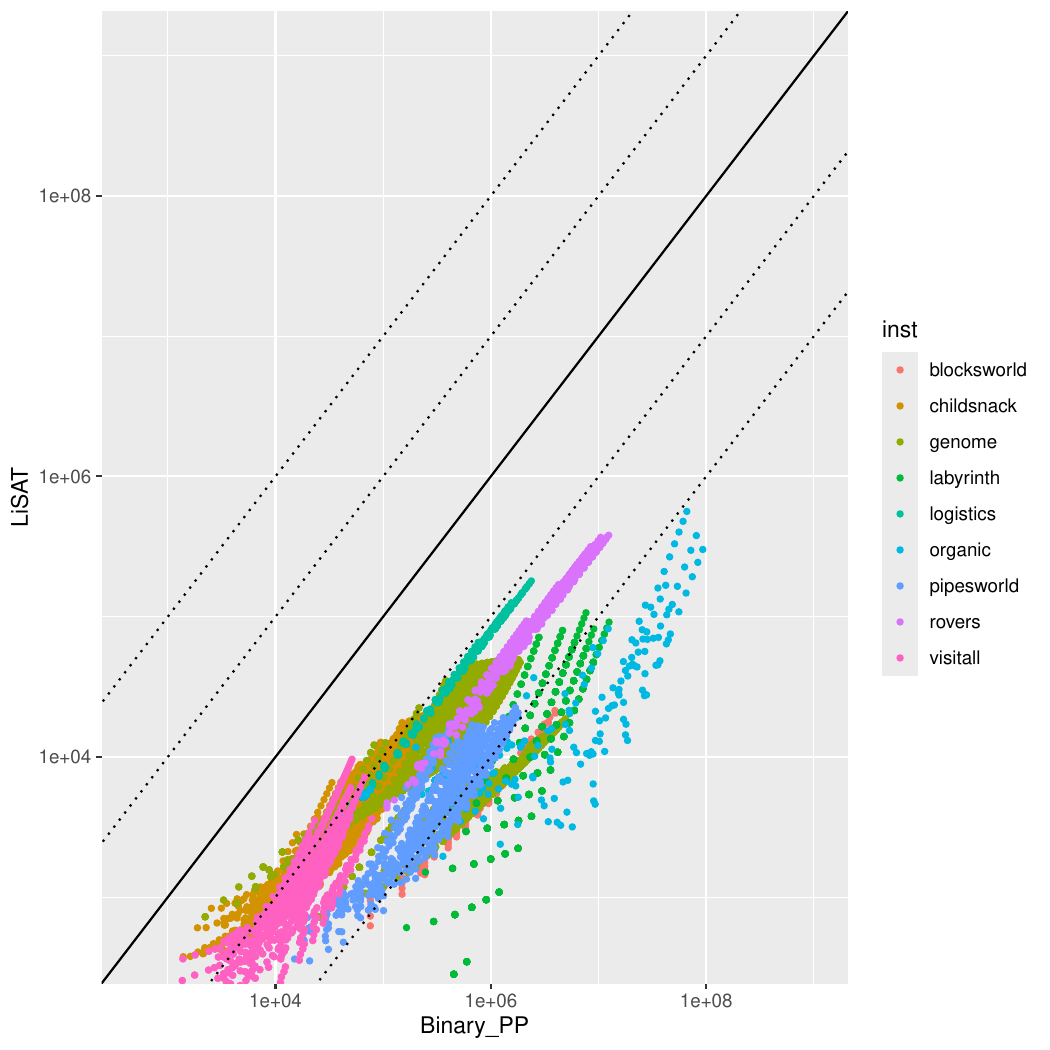}
        \caption{Number of clauses (log-log)}
    \end{subfigure}
    \hfill
    \begin{subfigure}[b]{0.32\textwidth}
        \includegraphics[width=\linewidth, height=4.5cm]{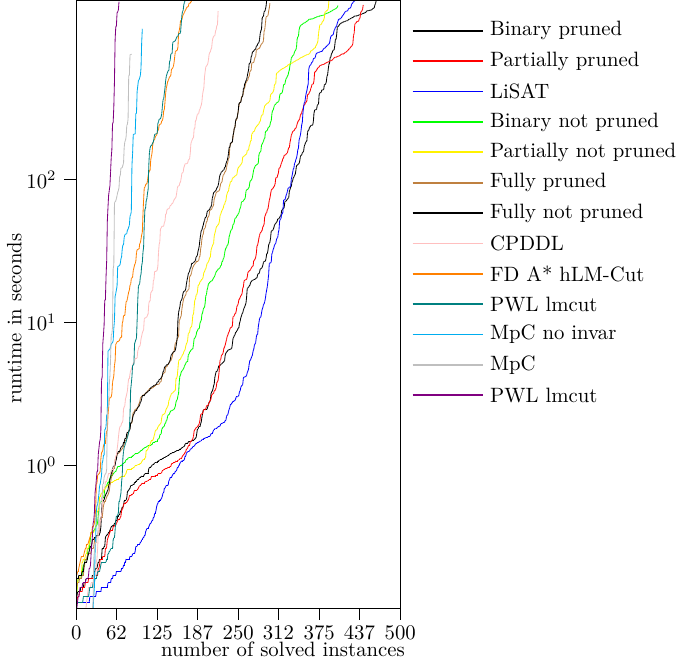}
        \caption{Coverage}
        \label{fig:covplot}
    \end{subfigure}
    \caption{Comparison of formula size, variable number and coverage}
    \label{fig:clauses}
\end{figure*}

The experiments were conducted on an AMD EPYC 9654 with a memory limit of 8GB and a time limit of 1800 seconds. A snapshot of the data and code used is archived at~\cite{zenodosnapshot}. The most up-to-date version of the codebase is maintained in a public repository\footnote{https://github.com/galvusdamor/PGSAT}. The results were obtained from the standard benchmark set for lifted planning\footnote{https://github.com/abcorrea/htg-domains} and validated with VAL\footnote{https://github.com/KCL-Planning/VAL}.
These benchmarks do not contain action costs, i.e., in our evaluation length-optimal is the same as cost-optimal.
Tab.~\ref{tab:results} presents the results of running multiple planners in optimal (left side) and satisficing (right side) modes. 
Fig.~\ref{fig:covplot} shows a cactus plot for the optimal mode.
Our encodings were ran with and without predicate pruning (PP) and use the Kissat SAT solver~\cite{BiereFallerFazekasFleuryFroleyksPollitt-SAT-Competition-2024-solvers}, the same solver employed by LiSAT. 
For the satisficing mode, we used the iteration algorithm proposed by ~\citeauthor{hoeller-behnke-icaps2022}~\shortcite{hoeller-behnke-icaps2022}: attempt to solve formulas for the bounds 10, 25, 50, 100, and 200 iteratively.
Each bound is allocated a time budget of $\frac{remaining\_run\_time}{|remaining\_bounds|}$, after which the next bound is attempted.

\ifextendedversion
\begin{figure*}[t] 
    \centering
    \begin{subfigure}[b]{0.48\textwidth}
        \includegraphics[width=\linewidth]{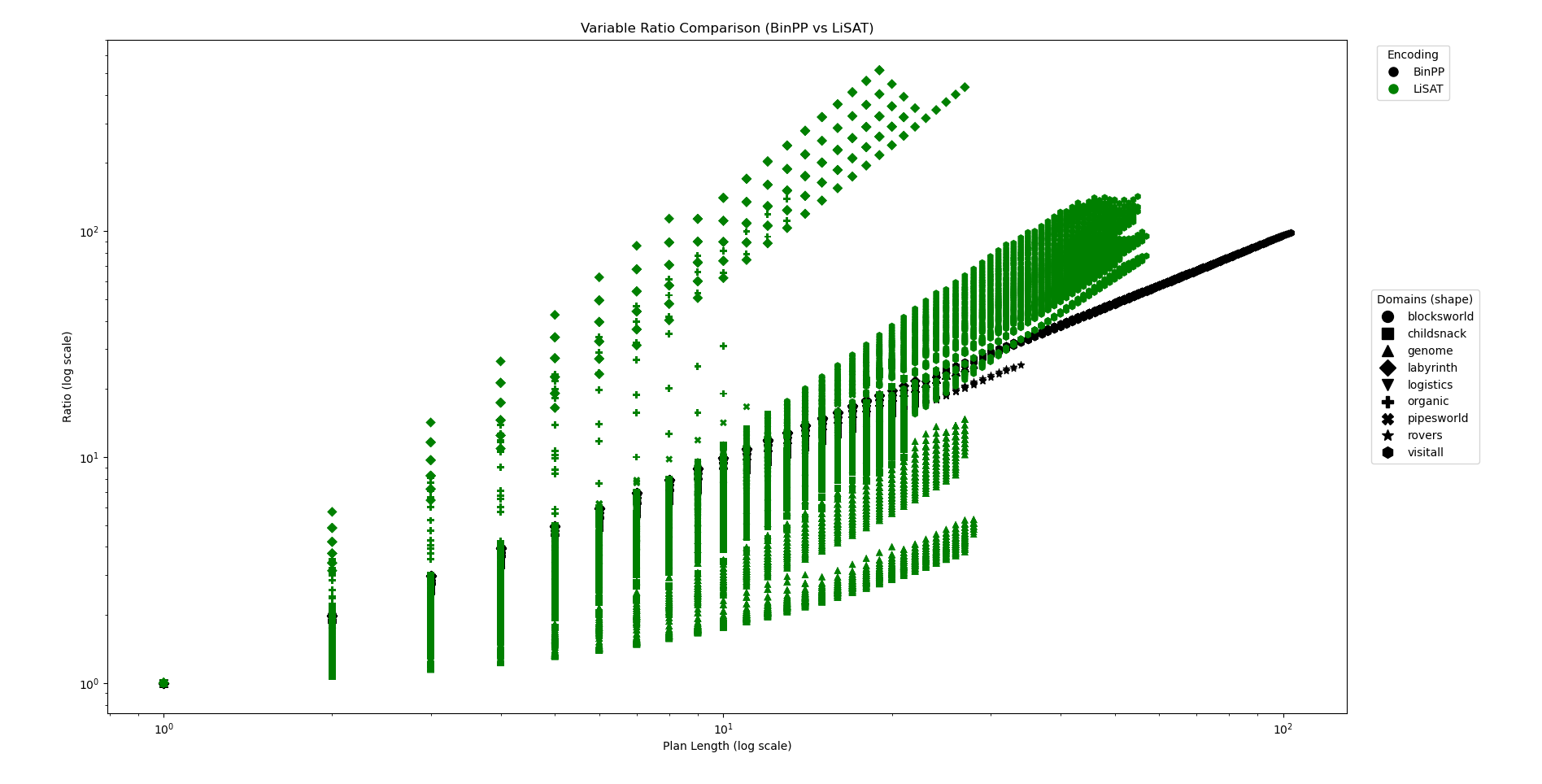}
        \caption{Variable Ratio Comparison (log/log)}
    \end{subfigure}
    \hfill
    \begin{subfigure}[b]{0.48\textwidth}
        \includegraphics[width=\linewidth]{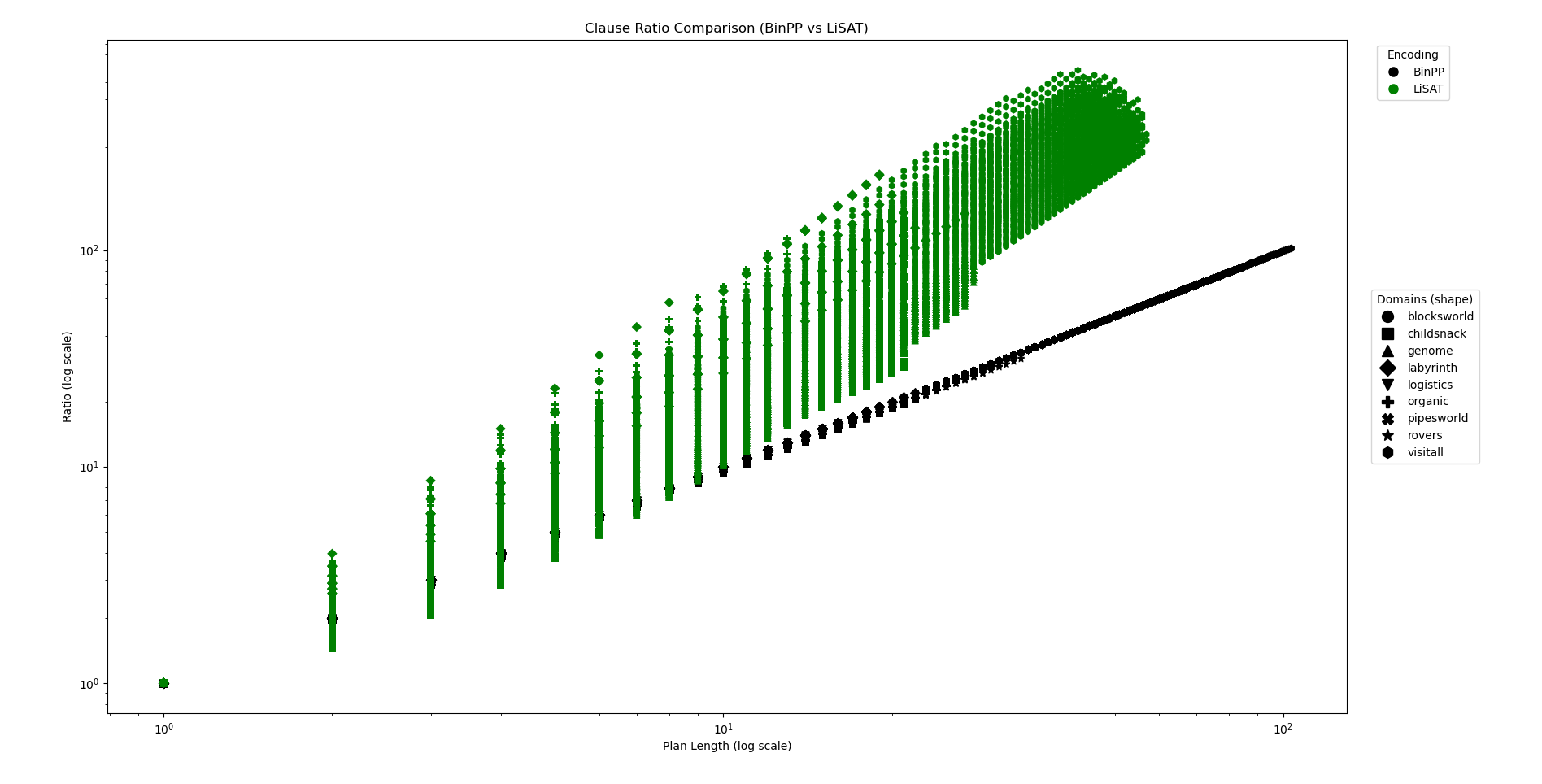}
        \caption{Clause Ratio Comparison (log/log)}
    \end{subfigure}
    \caption{Ration comparisons of number of variables/clauses with respect to a formula of length 1}
    \label{fig:ratios}
\end{figure*}

\begin{figure}
    \centering
    \includegraphics[width=\linewidth]{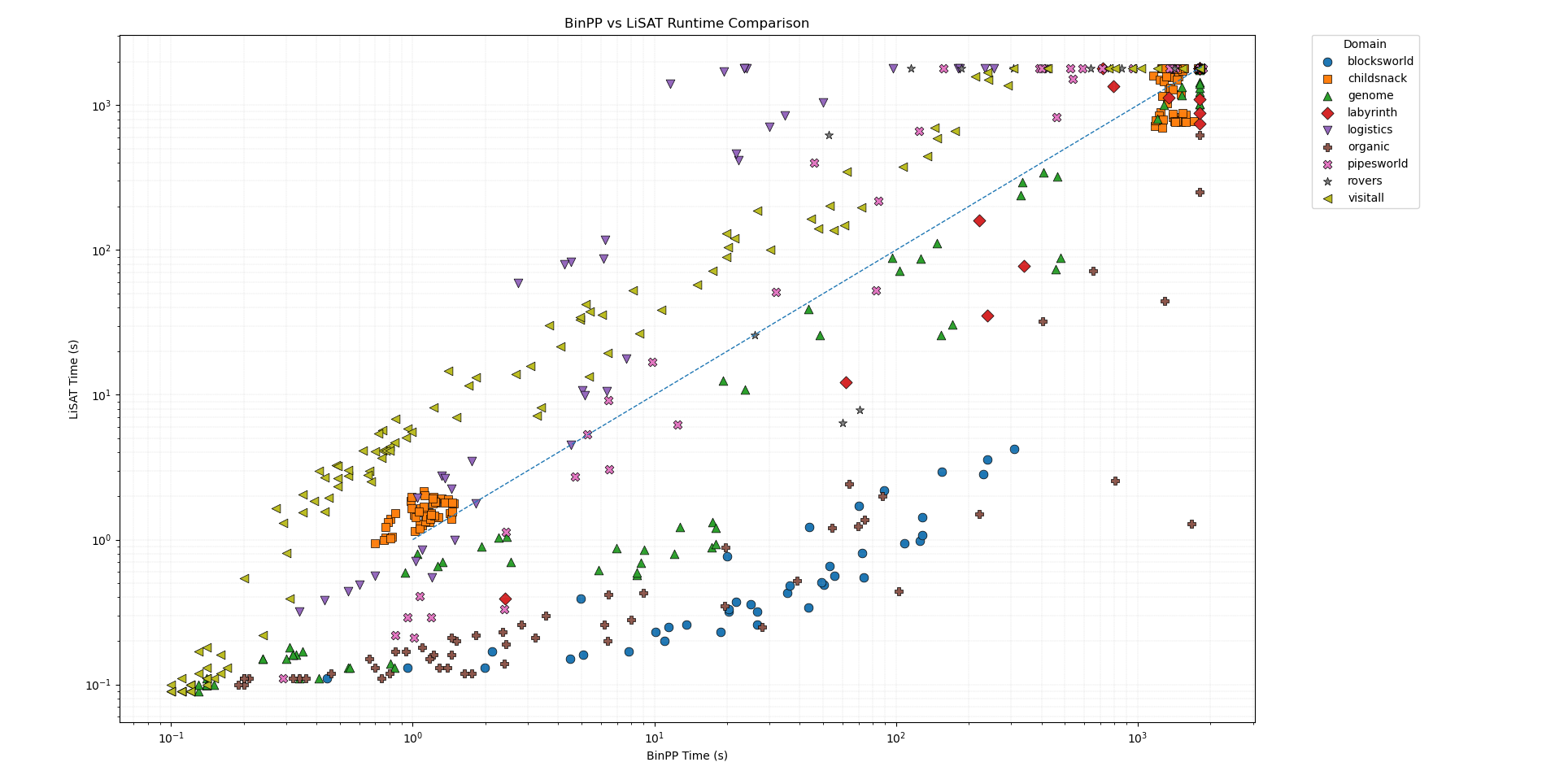}
    \caption{Solve Time Scatter Plot}
    \label{fig:timeplot}
\end{figure}
\fi

We compare against state-of-the-art lifted planners, Powerlifted (PWL) and CPDDL, as well as grounded planners Madagascar (MpC) and Fast Downward (FD). 
For PWL, we used A\textsuperscript{*} search with the $h^{max}$ heuristic (optimal mode), and alt-bfws1 and $h^{\textit{ff}}$ (satisficing mode).
We also used the optimal configuration (PWL A\textsuperscript{*} lmc) from \citeauthor{wichlacz-et-al-ecai2023}~\shortcite{wichlacz-et-al-ecai2023} presenting a lifted version of the lm-cut (lmc) heuristic~\cite{helmert-domshlak-icaps2009}, but could not fully reproduce their results even with technical support from an author; nevertheless, we include them for reference.
We use CPDDL with the best optimal and satisficing configuration reported in \citeauthor{horcik-fiser-ecai2023}~\shortcite{horcik-fiser-ecai2023} (Gaifman graphs with 95\% and 25\% reduction, A\textsuperscript{*} with $h^{lm\text{-}cut}$ and GBFS with $h^{\textit{ff}}$, respectively).
For reference, we included two grounded planners: Madagascar (state of the art grounded SAT-based planner, with and without invariant computation; no parallelism and algorithm S in optimal mode) and Fast Downward~\cite{helmert-jair2006} (with and without action splitting~\cite{elahi2024optimizing}) using A\textsuperscript{*} search and the lm-cut heuristic~\cite{helmert-domshlak-icaps2009} (in optimal mode) and the LAMA configuration (in satisficing mode). 
Coverage is the total number of instances solved across all domains, while the score is the sum of per-domain percentages of solved instances.

We do not compare against the approach by \citeauthor{robinson-et-al-icaps2009}~\shortcite{robinson-et-al-icaps2009}, as it is not meant for optimal planning (it allows for action parallelism) and no implementation is publicly available, nor against the QBF-based planner by \citeauthor{shaik-vandepol-icaps2022}~\shortcite{shaik-vandepol-icaps2022}, as it was outperformed by LiSAT.

\paragraph{Results -- Optimal}
In the optimal setting, search-based lifted methods are clearly outperformed by SAT-based methods: even the best approach (CPDDL A\textsuperscript{*} lmc) performs worse than our worst SAT-based approach (Fully, no PP).


The benefits of predicate pruning are most evident in Visitall. Although only one predicate (visited) is pruned, this alone eliminates approximately half of the grounded facts. Since this predicate is not part of any PLMG, all its facts require (expensive) explicit encoding, while all other non-static facts (of the at-robot predicate) form a single PLMG.
Here, predicate pruning substantially reduces the number facts that are not encoded compactly using PLMGs but explicitly (only the goal remains).
In Rovers 3 unary predicates (all of the type communicated\_X\_data) are pruned. The number of removed facts is not as significant, thus explaining the lack of improvement.
For Pipesworld (Binary), the difference observed is not caused by predicate pruning. The discrepancy is due to one encoding failing to find a satisfying assignment within the time limit.

Three of our encodings achieve a higher score than LiSAT.
Two of them also have higher coverage.
Furthermore, they fall short of LiSAT's performance in only three domains (GED, OS, Labyrinth), yet they remain competitive in two of them (GED, OS), with the difference in solved instances being $<$ 5\% of the total instances in the domain.
In contrast, of the five domains where our encodings (Binary PP) outperform LiSAT, four exhibit a $\geq 9\%$ increase in solved instances, with notable improvements in Logistics, Pipesworld, and Rover, where the difference is $\geq 20\%$.

\paragraph{Results -- Satisficing}
For brevity, we only report our best encoding (Binary, PP) here -- which is the second best in terms of score, outperformed only by PWL.
Although we underperform in terms of coverage compared to FD, LiSAT, and PWL, this is primarily due to GED, whose instance count skews coverage.
Importantly, our planner (and LiSAT) outperforms other lifted and grounded satisficing planners in the domain Blocksworld, Childsnack, Visitall, Labyrinth, and OS.
This indicates that SAT-based planners (ours and LiSAT) provided complementary capabilities to current search-based lifted planners, which would, e.g., be useful as part of a portfolio planner.


\paragraph{Results -- Formula Size}
Figure~\ref{fig:clauses} compares number of clauses and variable between Binary PP and LiSAT.
Interestingly, Binary PP always generates more clauses than LiSAT for the same plan length.
Even in the best-case (Childsnack) our encoding produces only less than one order of magnitude more clauses than LiSAT, while in most cases the difference is one to two orders of magnitude.
This is due to the increased size of the encoding needed for correctly encoding PLMG's semantics.
On the other hand, Binary PP requires fewer variables in most instances, sometimes up to two orders of magnitude.
The trend that our encoding scales linearly, while LiSAT's scales quadratically is however visible from LiSAT’s more rapid growth in number of variables and clauses.
For the number of clauses this is particularly evident in OS (light blue). We provide additional plots showing  the growth of the formulas more clearly in the appendix.

\ifextendedversion
Figure~\ref{fig:ratios} presents log-log plots comparing the growth of the number of variables and the number of clauses used by a formula of any length to the formula of length 1 between the predicate-pruned Binary encoding and LiSAT.
For this, we have taken for every planning problem, the number of clauses (variables, respectively) for encoding with plan length one: $N_1$.
Then for any higher plan length, we divided the number of clauses (variables, resp.) by $N_1$.
With this, we normalised the growth of the formula w.r.t.\ the formula for plan length 1.
This enables us to more easily compare instances of different size (number of objects) and across domains (different number and arity of predicates and actions).

In terms of both the number of clauses and the number of variables, we can see that our encoding exhibits a linear growth with the plan length while LiSAT exhibits a quadratic growth, as was expected. Additionally, it can also be seen that in LiSAT's formula, the growth is also dependent on the instance of each domain with harder instances exhibiting a bigger growth, while our encoding exhibits very little variance in the scaling across the domains.

\paragraph{Runtime}
Figure~\ref{fig:timeplot} presents a scatter plot of the solve time of both planners.
Interestingly, there is no clear dominance between LisSAT and Binary PP, but a dependence on the domain (e.g.\ Visitall for Binary PP and blocksworld for LiSAT).
This suggests that both approaches exhibit orthogonal capabilities.
\fi

\section{Conclusion}
We presented three encodings of partially lifted planning into SAT—keeping actions fully lifted while partially grounding the state description.
Unlike previous encodings whose formula size scaled quadratically with plan length, ours scales linearly.
Empirically, we have shown that our partially grounded and binary encodings outperform the state of the art in length-optimal planning  -- LiSAT -- and is competitive with it satisfying planning.
Further our new encodings provide complementary capabilities to current state-of-the-art lifted satisficing planners as they solve problem that are out of reach for search-based planners.

While our encodings target sequential plans, future work could explore integrating action parallelism. This may improve scalability and broaden the applicability of partially grounded encodings, especially in satisficing settings. The encodings may also be extended to support negative preconditions and conditional effects.

\section{Acknowledgements}
This publication is part of the project “Exploiting Problem Structures in SAT-based Planning” with file number OCENW.M.22.050 of the research programme Open Competition which is financed by the Dutch Research Council (NWO).

\bibliography{abbrv,extra,aaai2026,crossref}

\end{document}